\documentclass[sigconf]{acmart}

\usepackage{url}
\usepackage{cite}
\usepackage{algorithmic}
\usepackage{booktabs}
\usepackage{float}
\usepackage{graphicx}
\usepackage{multirow}
\usepackage{marvosym}
\usepackage{makecell}
\usepackage{microtype}
\usepackage{pifont}
\usepackage{subcaption}
\usepackage{textcomp}
\usepackage{wrapfig}
\usepackage{enumitem}
\usepackage{amsmath,amsfonts}
\usepackage{xcolor}
\usepackage{algorithm}
\usepackage[normalem]{ulem}
\usepackage{bm}
\newtheorem{definition}{Definition}

\usepackage{tikz}
\usepackage{svg}
\usepackage{utfsym}

\AtBeginDocument{%
  }

\setcopyright{acmlicensed}
\copyrightyear{2018}
\acmYear{2018}
\acmDOI{XXXXXXX.XXXXXXX}
\acmConference[Conference acronym 'XX]{Make sure to enter the correct
  conference title from your rights confirmation email}{June 03--05,
  2018}{Woodstock, NY}
\acmISBN{978-1-4503-XXXX-X/2018/06}

\begin{document}

\title{FairDRL-ST: Disentangled Representation Learning for Fair Spatio-Temporal Mobility Prediction}


\author{Sichen Zhao}
\affiliation{%
  \institution{RMIT University}
  \city{Melbourne}
  \country{Australia}}
\email{s3802901@student.rmit.edu.au}

\author{Wei Shao}
\affiliation{%
	\institution{CSIRO}
	\city{Melbourne}
	\country{Australia}}
\email{wei.shao@data61.csiro.au}

\author{Jeffrey Chan}
\affiliation{%
	\institution{RMIT University}
	\city{Melbourne}
	\country{Australia}}
\email{jeffrey.chan@rmit.edu.au}

\author{Ziqi Xu}
\affiliation{%
	\institution{RMIT University}
	\city{Melbourne}
	\country{Australia}}
\email{ziqi.xu@rmit.edu.au}

\author{Flora Salim}
\affiliation{%
	\institution{University of New South Wales}
	\city{Sydney}
	\country{Australia}}
\email{flora.salim@unsw.edu.au}

\renewcommand{\shortauthors}{Sichen et al.}

\begin{abstract}
	As deep spatio-temporal neural networks are increasingly utilised in urban computing contexts, the deployment of such methods can have a direct impact on users of critical urban infrastructure, such as public transport, emergency services, and traffic management systems. While many spatio-temporal methods focus on improving accuracy, fairness has recently gained attention due to growing evidence that biased predictions in spatio-temporal applications can disproportionately disadvantage certain demographic or geographic groups, thereby reinforcing existing socioeconomic inequalities and undermining the ethical deployment of AI in public services. In this paper, we propose a novel framework, FairDRL-ST, based on disentangled representation learning, to address fairness concerns in spatio-temporal prediction, with a particular focus on mobility demand forecasting. By leveraging adversarial learning and disentangled representation learning, our framework learns to separate attributes that contain sensitive information. Unlike existing methods that enforce fairness through supervised learning, which may lead to overcompensation and degraded performance, our framework achieves fairness in an unsupervised manner with minimal performance loss. We apply our framework to real-world urban mobility datasets and demonstrate its ability to close fairness gaps while delivering competitive predictive performance compared to state-of-the-art fairness-aware methods.
\end{abstract}

\begin{CCSXML}
	<ccs2012>
	<concept>
	<concept_id>10010147.10010257.10010293</concept_id>
	<concept_desc>Computing methodologies~Machine learning approaches</concept_desc>
	<concept_significance>500</concept_significance>
	</concept>
	<concept>
	<concept_id>10010147.10010257.10010293.10010319</concept_id>
	<concept_desc>Computing methodologies~Learning latent representations</concept_desc>
	<concept_significance>500</concept_significance>
	</concept>
	<concept>
	<concept_id>10002951.10003227.10003236.10003237</concept_id>
	<concept_desc>Information systems~Geographic information systems</concept_desc>
	<concept_significance>500</concept_significance>
	</concept>
	<concept>
	<concept_id>10003456.10003457.10003567.10010990</concept_id>
	<concept_desc>Social and professional topics~Socio-technical systems</concept_desc>
	<concept_significance>500</concept_significance>
	</concept>
	</ccs2012>
\end{CCSXML}

\ccsdesc[500]{Computing methodologies~Machine learning approaches}
\ccsdesc[500]{Computing methodologies~Learning latent representations}
\ccsdesc[500]{Information systems~Geographic information systems}
\ccsdesc[500]{Social and professional topics~Socio-technical systems}

\keywords{Spatio-temporal prediction, Urban mobility prediction, Representation learning for geospatial data}


\maketitle

\section{Introduction}
Predicting the spatio-temporal dynamics of urban areas has become an increasingly important application of representation learning, driven by the proliferation of the Internet of Things in cities. Deep spatio-temporal neural networks are applied in various domains, including predicting crowd behaviour, allocating mobility resources, and detecting air pollution. In the design of spatio-temporal methods, accuracy is a critical factor, as resource misallocation can lead to significant economic and productivity losses. Many researchers have sought to improve spatio-temporal prediction performance by incorporating additional exogenous contextual datasets or enhancing the extraction of spatio-temporal dynamics \citep{taxibj, wang2018cross, zonoozi2018periodic, lin2019deepstn+, yao2019revisiting, li2019densely, shao2024spatio}.

While the accuracy of deep learning methods has improved with the growing availability of multi-sensory data, fairness considerations have recently gained significant attention in both research and practice. Designers of spatio-temporal methods must ensure that their methods do not have unintended social implications, as their predictions can impact people's access to economic and social opportunities. Deep spatio-temporal models have been shown to produce biased predictions that reinforce socioeconomic inequality when input data are not properly handled. For example, housing price data may reveal discrimination embedded in urban development, while transportation data may expose resource distribution patterns that favour wealthier areas. Without appropriate pre-processing, these biases can easily be introduced during the training phase, leading to unfair methods that could negatively affect the broader socioeconomic system.

 \begin{figure}[t]
	 \begin{center}
		     \includegraphics[width=0.45\textwidth]{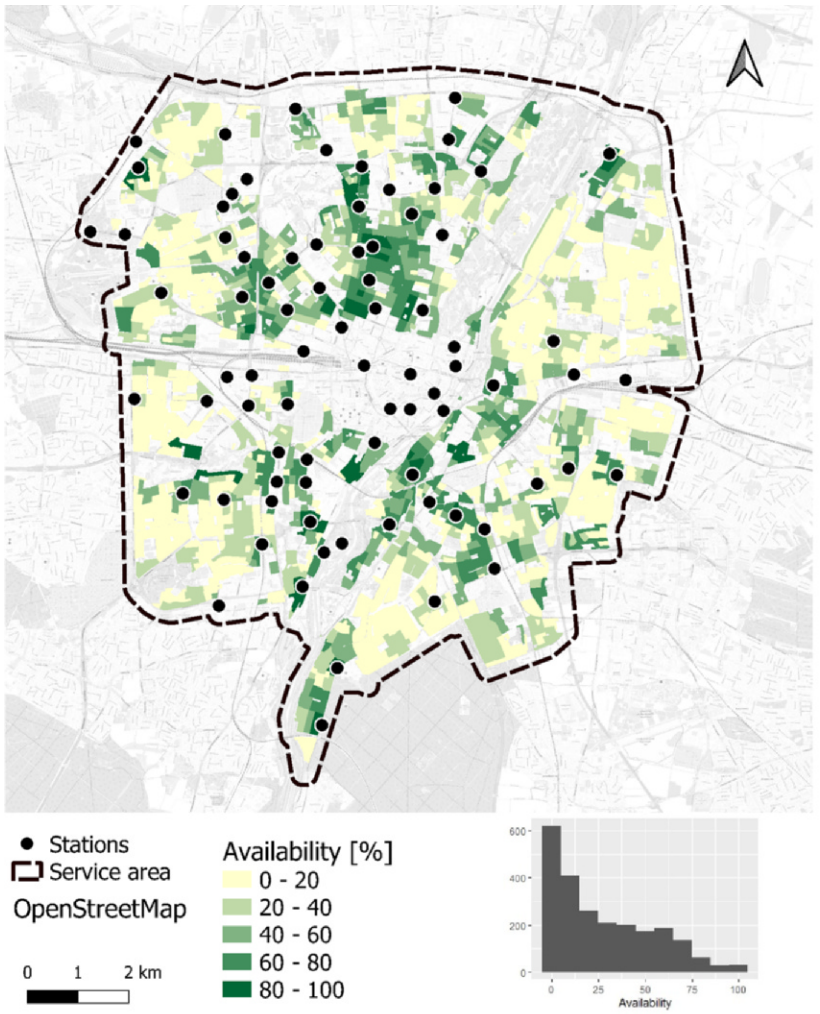}
		 \end{center}
	 \caption{The location and availability of bike-sharing stations per residential block in Munich.}
	 \label{fig:illu}
\end{figure}

When biases that are often correlated with sensitive attributes such as income or race are not separated from the input data, a harmful feedback loop can occur. For example, the underestimation of resource demand in disadvantaged communities may lead to reduced service provision, further limiting their access and reinforcing inequality. A real-world example of this issue can be seen in the distribution of bike-sharing station availability in Munich, as shown in Figure~\ref{fig:illu}. The map reveals a clear spatial disparity: central and affluent areas have significantly higher station availability, while peripheral or less privileged neighbourhoods experience limited access, highlighting a potential source of structural bias in mobility resource allocation. Therefore, identifying and removing the influence of sensitive attributes from the data is essential when constructing fair and socially responsible spatio-temporal prediction methods.

Disentangled Representation Learning (DRL) involves extracting and learning explicit, interpretable low-dimensional representations that capture the underlying factors of variation in data. In conventional machine learning methods, representations often become entangled, where multiple factors of variation are combined into a single representation. DRL aims to disentangle these underlying factors, including both relevant and sensitive attributes. By explicitly modelling and separating such attributes, patterns linked to sensitive attributes can be isolated and treated separately. This approach allows the development of prediction models that account for fairness and mitigate biases. Incorporating DRL into spatio-temporal prediction models helps ensure these models do not unintentionally reinforce socioeconomic inequalities or discriminate against specific groups. By identifying and eliminating biases related to sensitive attributes, DRL prevents unfair distribution of resources or access to services.

In addition to promoting fairness, incorporating DRL into spatio-temporal prediction methods offers several additional advantages. Firstly, it enhances interpretability by isolating and disentangling factors of variation in the data. This allows researchers to better understand and explain the relationships between different attributes and their influence on predictions \citep{zhao2022measuring}. Moreover, DRL can assist in generalisation across different contexts or domains. By segregating relevant and sensitive attributes, the method can concentrate on discovering the underlying dynamics of the spatio-temporal data without being affected by biased or discriminatory factors. Consequently, the learned representations can be applied to a variety of tasks.

Systematic socioeconomic and racial discrimination is often embedded in urban datasets, contaminating the training signals used by predictive models. For instance, police incident records are frequently used to predict crime patterns, but they only reflect reported crimes rather than actual crime occurrences \citep{wu2021fair, mason2019reporting}, introducing significant bias. Similarly, housing prices or service accessibility often reflect structural inequalities rather than purely behavioural data. Previous research has attempted to mitigate such biases by removing or suppressing sensitive information under supervision \citep{yan2020fairness, yan2021equitensors}. In contrast, we introduce a DRL mechanism that separates sensitive attributes, such as race, income, or location-specific biases, from the learned representations in an unsupervised manner. To improve the model’s ability to isolate these latent factors, we incorporate additional regularisation strategies that help guide the separation of sensitive signals from those relevant to prediction.

Although fairness has become an important topic in machine learning, it remains under-explored in spatio-temporal prediction tasks. Existing methods often enforce fairness constraints within the objective function, relying on supervised labels or predefined fairness metrics. These approaches are difficult to generalise across domains and may lead to overcompensation, which in turn degrades prediction performance \citep{yan2020fairness, kang2024promoting}. Our framework does not enforce fairness metrics directly during training. Instead, it achieves fairness by learning unbiased representations through disentanglement. This unsupervised approach eliminates the need for sensitive group labels or fairness-specific annotations, allowing the model to generalise more effectively and avoid the drawbacks associated with supervised fairness regularisation.

Our main contributions are summarised as follows:
\begin{itemize}
	\item We propose FairDRL-ST, a fairness-aware spatio-temporal prediction model for mobility resource demand. FairDRL-ST captures non-sensitive temporal and spatial dependencies while effectively incorporating multi-dimensional exogenous attributes to enhance forecasting performance.
	
	\item We develop a technically innovative unsupervised fairness framework by integrating adversarial learning with disentangled representation learning. FairDRL-ST introduces a novel regularisation strategy that encourages the separation of sensitive and task-relevant factors in the latent space, enabling fair representation learning without relying on demographic group labels.
	
	\item We conduct extensive experiments on real-world urban mobility datasets. Results demonstrate that FairDRL-ST consistently reduces fairness disparities across demographic and regional groups while maintaining or improving predictive performance compared to state-of-the-art fairness-aware baselines.
\end{itemize}

The remainder of this paper is organised as follows. Section~\ref{Related} reviews the literature on fairness in spatio-temporal prediction and representation learning. Section~\ref{Preliminaries} presents key definitions relevant to our framework. In Section~\ref{Methodology}, we detail the proposed FairDRL-ST framework, including its architectural components and training strategy. Section~\ref{Experiments} describes the experimental setup, datasets, and evaluation metrics, followed by a comprehensive analysis of the results. Finally, Section~\ref{Conclusion} concludes the paper and outlines potential directions for future work.

\section{Related Work}
\label{Related}
Recent works have already recognised the challenges in making fairness-aware predictions with deep learning models trained on a large volume of data \citep{chirigati2016data}; our focus is on creating a general approach that can be directly used in various mobility prediction tasks. Although the machine learning community has made progress on mobility prediction \citep{taxibj, lin2019deepstn+, zonoozi2018periodic, li2019densely, yao2019revisiting}, representation learning \citep{fabius2014variational, zhu2020s3vae, li2018disentangled,xu2023disentangled,xu2023disentangl} and reducing discriminatory predictions \citep{ekstrand2018privacy, zemel2013learning, yan2021equitensors,xu2022assessing}, the sub-area of how to generate fairness-aware predictions for spatio-temporal data still lacks attention.

\subsection{Mobility Prediction}
It is crucial to consider urban density and human mobility to comprehend and define urban environments. For example, population density is the foundation for designating urban regions and quantifying the amount of urbanisation. Thanks to the IoTs, we can collect mobility data with great precision and at a very low cost, which gives us accurate estimates of the population dynamics and fine-grained urban density in urban areas. However, the question of how to use these enormous amounts of data and generate precise predictions is yet unanswered.

To accurately anticipate traffic, previous efforts have concentrated on capturing the spatio-temporal dependency hidden beneath the traffic data. In addition to more conventional techniques like Seasonal ARIMA prediction \citep{moreira2013predicting}, deep learning techniques are being included and used in an increasing number of studies for mobility forecasting. It is suggested using ST-ResNet \citep{taxibj} to capture spatial dependencies via a stack of residual convolution layers. To capture temporal dynamics, it also stacks the frames from two distinct periods. The outcomes show that the goal of employing residual units to address the vanishing gradient problem is to improve the accuracy of mobility predictions by capturing distant spatial information. To more thoroughly examine the effectiveness of temporal features, recurrent-based algorithms were developed to capture temporal correlations. PCRN \citep{zonoozi2018periodic} is proposed to utilise a convolutional recurrent network (CRN) to extract entangled spatio-temporal representations before leveraging the CRN's hidden state to update the periodic representations. 

While the aforementioned methods show that capturing spatial and temporal correlations enhances prediction accuracy, our major purpose is to understand the limitations of representation learning when we relax our assumptions about the features and the primary goal of the target application and attempt to produce fairness-aware predictions by separating the biased features.

\subsection{Disentangled Representation Learning}
Variational Autoencoder (VAE), an unsupervised generative learning technique, is the foundation for the majority of earlier works on the disentangled representation learning (DRL) problem. When the new hyperparameter $beta$ is introduced to the inference model to impose an information bottleneck on the prior, $beta$-VAE \citep{betavae} compels the inference model to disentangle the latent representation. Further decomposing the objective function, FactorVAE \citep{factorVAE} attempts to improve disentanglement by penalising the overall correlation of the learned representation. To improve the effectiveness and identifiability of disentanglement, \citet{KhemakhemKMH20} propose the iVAE, which has a theoretical guarantee for the identification of the learned representations by using an additionally observed variable. \citet{XuCLL0Y24, cheng2024instrumental, 10791303, ChengXL0LL24} further apply this technique in causal inference in the presence of unobserved variables.

Regarding sequence modelling, a few earlier papers \citep{fabius2014variational, chung2015recurrent, bayer2014learning} have expanded VAE to video and audio data. To distinguish between static and dynamic elements of sequential data, S3VAE \citep{zhu2020s3vae} is proposed. \citep{li2018disentangled} also suggests a different method that focuses on separating the dynamic factors from the static factors.

\subsection{Fairness in Machine Learning}
There is a large body of research on fair machine learning, including work by \citep{chouldechova2018frontiers, ekstrand2018privacy, hardt2016equality, dwork2012fairness, zemel2013learning,XuKON25,LuoHY0024}. However, most of these studies do not focus on spatio-temporal applications. One notable exception is the work by \citet{yan2020fairness}, who propose a fairness-aware prediction framework for urban mobility by incorporating fairness as a regularisation term. This approach, however, is based on supervised learning and depends on the availability of demographic group labels.

In the unsupervised setting, several methods have been proposed for learning fair representations. Examples include variational inference approaches \citep{louizos2015variational}, disentangled representation learning techniques \citep{ruoss2020learning,xu2023disentangled}, and early encoder-based fairness models \citep{zemel2013learning}. Adversarial learning has also been widely adopted to mitigate bias by preventing sensitive information from being captured in the learned representations \citep{sadeghi2020imparting, wadsworth2018achieving, xu2019achieving, xu2019fairgan}. For example, \citet{madras2018learning} design an encoder-decoder structure where an adversarial module attempts to infer sensitive attributes, while the encoder is trained to minimise the adversary's success. Although effective in discarding sensitive information, these methods generally lack mechanisms for validating or interpreting the discarded components. In addition, very few studies address fairness in forecasting tasks that involve continuous and spatially distributed sensitive attributes, such as regional income maps.

To overcome these limitations, we propose a DRL based framework that explicitly separates sensitive and task-relevant factors. This design enables our framework to mitigate biases while enhancing fairness through improved adversarial training in spatio-temporal prediction.

\section{Preliminaries}
\label{Preliminaries}
In this section, we introduce key concepts relevant to our work. We begin by formally defining the mobility demand prediction task, including the spatial partitioning of urban areas and the temporal modelling of resource demand. We then provide an overview of Variational AutoEncoders (VAEs) and disentangled representation learning (DRL), which form the foundation of our methodology. Specifically, we explain how VAEs encode data into latent variables and how disentanglement is achieved through regularisation terms such as mutual information and total correlation.

\subsection{Mobility Demand Prediction}
In this section, we provide a concise overview of the mobility demand prediction task. The goal is to estimate the demand for mobility resources across different regions within a predefined urban area over time. To facilitate this, the study area is divided into a grid of spatial cells, each corresponding to a distinct region. We then define the spatial units, the distribution of mobility resources, and the prediction targets in the following formal definitions.

\begin{definition}[Region~\citep{taxibj}]
	We partition the study area $\mathcal{C} = \{ c_1, c_2, \dots, c_n \}$ into a grid map consisting of $n$ cells in total. Each cell represents a region, and the grid is split based on longitude and latitude. The population percentage of cell $c_i$ is represented as $p_i$, with $\sum_{i=1}^{n} p_i = 1$.
\end{definition}

\begin{definition}[Mobility resource distribution~\citep{yan2020fairness}]
	Let $x_{i,t}$ represent the available mobility resource for cell $c_i$ in the study area $\mathcal{C}$ at time interval $t$. The distribution of mobility resources at time interval $t$ for the study area $\mathcal{C}$ can be denoted as a tensor $X_t \in \mathbb{R}^{n}$. Consequently, the historical observations over a time period $T$ can be represented by $X_{1:T}$.
\end{definition}

\begin{definition}[The prediction of mobility resource demand\\~\citep{yan2020fairness}]
	Let $\hat{y}{i,t}$ and $y{i,t}$ represent the predicted demand and ground truth demand for region $c_i$ at time $t$, respectively. Additionally, let $E_{T}[\hat{y}_{i,t}]$ denote the average predicted value for cell $c_i$ in the study area $\mathcal{C}$ over the time period $T$.
\end{definition}

\subsection{Variational AutoEncoder and Disentangled Representation Learning}
The VAE is an important class of generative models introduced by \citet{vae}. It consists of an encoder and a decoder, which work together to learn a low-dimensional representation of the data. In this model, it is assumed that the data $x$ is generated by a random process, modelled as $p_{\theta}(z)p_{\theta}(x \vert z)$. 

The encoder, also known as the inference model, is a neural network component denoted as $q_{\phi}(x \vert z)$, which approximates the posterior distribution $p_{\theta}(x \vert z)$. The encoder maps input data $x$ to a latent space representation $z$, capturing the key factors of variation in the data.

The decoder, also known as the generative model, aims to reconstruct the original data given the latent representation $z$. It is characterised by the conditional distribution $p_{\theta}(x \vert z)$, where $\theta$ represents the parameters of the decoder. The decoder takes a latent code $z$, sampled from a prior distribution $p_{\theta}(z)$, and generates a reconstructed output $x$, allowing the model to generate new samples similar to the training data. The VAE model is trained by maximising the marginal likelihood, incorporating both the reconstruction loss and a regularisation term.
\begin{equation}
	log \ p_{\theta }( x) =D_{KL}( q_{\phi }( z\vert x) \vert \vert p_{\theta }( z\vert x)) +\mathcal{L}_{VAE}( \theta ,\phi ;x)
\end{equation}

The regularisation term encourages the encoder's approximate posterior $q_{\phi}(z \vert x)$ to match the prior distribution $p_{\theta}(z)$, promoting disentangled representations in the latent space. By jointly learning the encoder and decoder, VAEs enable efficient and probabilistically meaningful generation and manipulation of complex data. However, directly computing the marginal likelihood empirically is challenging due to the intractability of the posterior distribution. \citet{vae} introduce a substitute called the evidence lower bound (ELBO), which can be optimised using stochastic gradient descent:
\begin{equation}
	\mathcal{L}( \theta ,\phi ;x) =-D_{KL}( q_{\phi }( z\vert x) \vert \vert p_{\theta }(z\vert x)) +E_{q_{\phi }( z\vert x)}[ log\ p_{\theta }( x\vert z)]
	\label{eq:ELBO}
\end{equation}

By minimising the reconstruction loss (the second term in Equation \ref{eq:ELBO}), the model is encouraged to produce more accurate reconstructions of both the input and synthetic data.

It is first discovered by $\beta$-VAE \citep{betavae} that the first term in Equation \ref{eq:ELBO} can be used to impose constraints on the information bottleneck, encouraging the model to learn more disentangled representations. The first term in Equation \ref{eq:ELBO} is decomposed as follows:
\begin{equation}
	\begin{aligned}
		D_{KL} (q_{\phi } (z\vert x)\vert \vert p_{\theta } (z\vert x)) & =\sum _{d} D_{KL}( q_{\phi }( z_{j}) \vert \vert p( z_{j}))+I( x,z) \\
		& \quad\quad+D_{KL}\left( q( z) \vert \vert \prod _{j=1} q( z_{j})\right),
		\label{eq:factor_decomp}
	\end{aligned}
\end{equation}
where $z\in \mathbb{R}^{d}$ and the original ELBO is structured as:
\begin{equation}
	\begin{aligned}
		\mathcal{L} (\theta ,\phi ;x) & =E_{q_{\phi } (z\vert x)} [log\ p_{\theta } (x\vert z)]-\sum _{d} D_{KL}( q_{\phi }( z_{j}) \vert \vert p( z_{j}))\nonumber\\
		& \quad\quad-I( x,z) -D_{KL}\left( q( z) \vert \vert \prod _{j=1} q( z_{j})\right) 
		\label{eq:final_elbo}
	\end{aligned}
\end{equation}

The first term in Equation \ref{eq:factor_decomp} is defined as \textit{dimension-wise KL divergence}, which puts constraints on the generated latent representation $z$ and pushes them towards their predefined Gaussian prior \citep{li2020pri}. The second term is called the \textit{mutual information}, which indicates the amount of information maintained by the learned representations about its input. DRL has been extensively utilised for tasks such as image generation, text generation, and dimensionality reduction, providing an effective tool for unsupervised learning and representation learning.

\begin{figure*}[t]
	\begin{center}
		\includegraphics[width=0.8\textwidth]{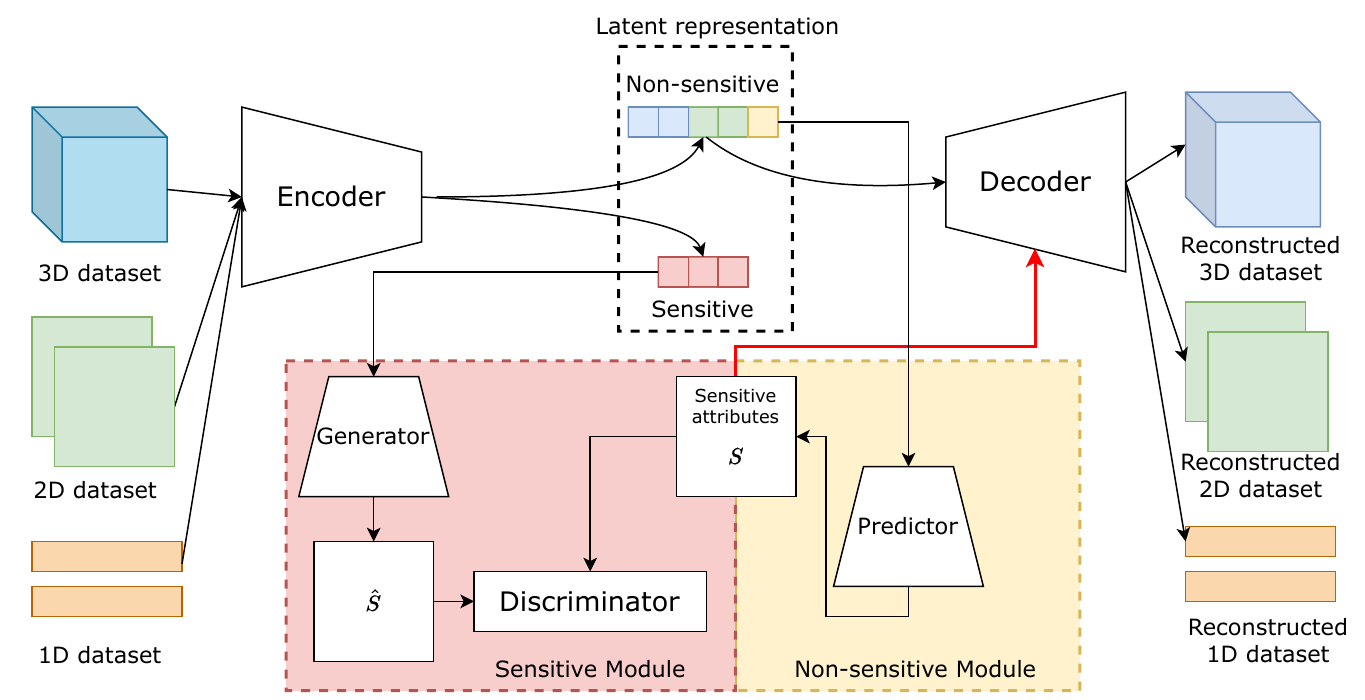}
	\end{center}
	\caption{The overall architecture of our proposed FairDRL-ST. The encoder and decoder are trained using disentangled representation learning to separate the representation into sensitive and non-sensitive components. The non-sensitive features are fed into the decoder to minimize reconstruction error, while the sensitive features are passed to an adversarial module. A discriminator attempts to distinguish the true sensitive attribute $S$ from the generated one $\hat{S}$ produced by the generator.}
	\label{fig:arch}
\end{figure*}

\section{Methodology}
\label{Methodology}
This section introduces our proposed framework, FairDRL-ST, which aims to learn disentangled representations that separate sensitive and non-sensitive information in an unsupervised manner. We begin with an overview of the framework, followed by detailed descriptions of the data pre-processing process, disentanglement strategies, generative modelling approach, and fairness evaluation metrics.

\subsection{Framework Overview}
\label{sec:framework_overview}
Figure~\ref{fig:arch} presents the overall architecture of our proposed framework, which consists of four main components: an encoder, a sensitive module, a non-sensitive module, and a decoder. The framework is designed to learn fair spatio-temporal representations by disentangling sensitive and non-sensitive factors in an unsupervised manner.

\begin{itemize} [leftmargin=0.5cm]
	\item \textit{Encoder}: This module takes input datasets with different dimensions and extracts latent representations from them.
	\item \textit{Sensitive Module}: Uses an adversarial approach that forces the model to separate the features relevant to the sensitive demographic attribute into the sensitive group.
	\item \textit{Non-sensitive Module}: Helps remove information relevant to the sensitive demographic attribute from the non-sensitive group.
	\item \textit{Decoder}: Reconstructs the input datasets using the non-sensitive attributes and the ground truth of sensitive attributes.
\end{itemize}

Each component of the framework is elaborated in the subsequent sections. The \textit{Encoder} and its data transformation process are introduced in Section~\ref{sec:data_preprocessing}, which details our input preparation and feature extraction pipeline. The disentanglement mechanisms of the \textit{Sensitive Module} and \textit{Non-sensitive Module} are jointly discussed in Section~\ref{sec:disentanglement}, where we describe the adversarial learning setup. The role of the \textit{Decoder}, together with the overall generative modelling strategy, is presented in Section~\ref{sec:vae}, followed by the final loss formulation and training objective in Section~\ref{sec:final_objective}.

\subsection{Data Pre-processing and Feature Extraction}
\label{sec:data_preprocessing}

Given the heterogeneous nature of the input datasets, which include 1D, 2D, and 3D formats, we first transform them into a unified rasterised structure before feeding them into the encoder. Following the approach of \citet{yan2021equitensors}, all datasets are converted into a grid-based representation with non-overlapping spatial cells. Missing values are imputed using local averages, and all features are rescaled to the range $[0, 1]$ via max scaling to facilitate efficient model training. Datasets sharing the same dimensionality undergo a consistent pre-processing strategy.

For {1D datasets}, such as temperature, which vary over time but not space, we aggregate the time series into 30-minute intervals and concatenate them into a tensor of shape $T \times C_1$, where $T$ is the number of time steps and $C_1$ is the number of 1D variables.

For {2D datasets}, which capture static spatial attributes like land use or infrastructure layouts, we rasterise point and line features by counting their occurrences in each grid cell or allocating attribute values proportionally by area. The resulting tensor has shape $H \times W \times C_2$, where $H$ and $W$ denote the grid height and width.

For {3D datasets}, such as dynamic mobility traces, we aggregate the spatio-temporal signals into tensors of shape $H \times W \times T \times C_3$.

\begin{figure*}[t]
	\begin{center}
		\includegraphics[width=0.75\textwidth]{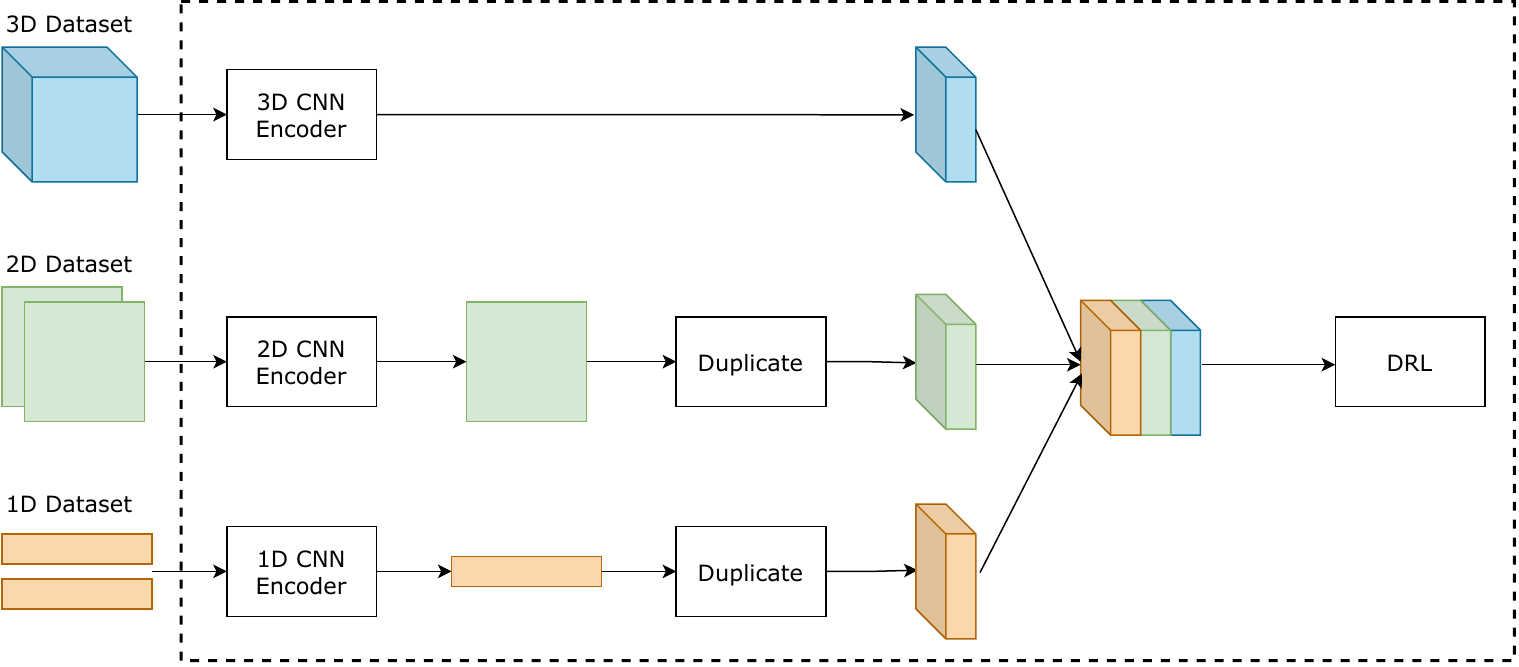}
	\end{center}
	\caption{The detailed structure of the encoder. We adopt the Convolutional Denoising Autoencoder (CDAE)~\citep{yan2021equitensors}, as our input data is similar in format to that used in the original study. It includes 1D, 2D, and 3D data.}
	\label{fig:encoder}
\end{figure*}

Following pre-processing, each dataset is passed to a dedicated encoder that maps it into a compact representation. As shown in Figure~\ref{fig:encoder}, each encoder consists of three convolutional layers with 32, 16, and 1 kernels, respectively. The resulting feature maps have dimensions of $H \times W \times T \times 1$ for 3D inputs, $H \times W \times 1$ for 2D inputs, and $T \times 1$ for 1D inputs.

To ensure alignment for downstream processing, we duplicate the 2D features along the temporal axis and the 1D features across spatial dimensions, producing unified representations of shape $H \times W \times T \times 1$. These tensors are stacked and passed into the DRL module.

The DRL module applies two additional 3D convolutional layers with 16 and 4 filters, respectively, to extract meaningful latent representations. These representations are then split into two distinct groups: one capturing sensitive information and the other encoding non-sensitive features. This separation serves as the basis for the fairness-aware mechanisms described in subsequent sections.

\subsection{Disentangling Sensitive Attributes}
\label{sec:disentanglement}

To promote fairness in the learned representations, we design a disentanglement mechanism that separates sensitive and non-sensitive attributes within the latent space. Specifically, we divide the latent representation $z$ into two components: $z^S$, which captures information associated with sensitive attributes $S$, and $z^{Ns}$, which encodes the remaining non-sensitive factors.

During reconstruction, the decoder leverages only $z^{Ns}$ in conjunction with the ground-truth sensitive attributes $S$ to reconstruct the input. By excluding $z^S$ from the reconstruction process, the model is encouraged to isolate sensitive information from the non-sensitive features, thereby supporting disentanglement.

However, without additional constraints, there is no guarantee that $z^S$ and $z^{Ns}$ are truly disentangled. To address this, we introduce two auxiliary regularisation strategies:

\begin{itemize}[leftmargin=0.5cm]
	\item \textbf{Adversarial Regularisation for $z^S$}: We employ a generator-discriminator pair. The generator takes $z^S$ as input and produces synthetic sensitive attributes $\hat{S}$, while the discriminator attempts to distinguish between the real attributes $S$ and the generated $\hat{S}$. This adversarial training ensures that $z^S$ retains sufficient information to reproduce $S$ accurately.
	
	\item \textbf{Predictive Regularisation for $z^{Ns}$}: We introduce a predictor that attempts to infer $S$ from $z^{Ns}$. To discourage leakage of sensitive information into the non-sensitive representation, we penalise successful predictions. The loss function for this module is:
	\begin{equation}
		\mathcal{L}_{Ns} = \frac{1}{n} \sum_{i=1}^{n} \left| P(z^{Ns}_i) - S_i \right|,
	\end{equation}
	where $P(z^{Ns}_i)$ denotes the predicted sensitive attribute for the $i$-th sample. This term is later multiplied by $-1$ in the final objective to penalise accurate predictions and enforce fairness through obfuscation.
\end{itemize}

Together, these two complementary regularisation components guide the model to explicitly separate sensitive signals from the rest of the latent space, ensuring that the non-sensitive representations are both informative and fair.

\subsection{Spatio-temporal Variational AutoEncoder}
\label{sec:vae}

We formally define $\mathcal{D} = \{X\}$ as a dataset that consists of multiple historical observation sequences $X_{1:T}$. In this work, we propose a spatio-temporal VAE architecture to learn disentangled representations from heterogeneous urban datasets. Our assumption is that each input $x$ is generated from a corresponding latent representation $z$, which can be separated into two disentangled subgroups: $z^{Ns}_{1:T}$ that contain non-sensitive information, and $z^{S}$ that contain sensitive information.

\paragraph{Prior.}
We define the prior of $z^{S}$ as a standard Gaussian distribution, $z^{S} \sim \mathcal{N}(0, 1)$. The time-varying $z^{Ns}_{1:T}$ follow a sequential prior:
\begin{equation}
	z_{t}^{Ns} |z_{< t}^{Ns} \sim \mathcal{N}\left( \mu _{t} ,\mathrm{diag}\left( \sigma _{t}^{2}\right)\right),
\end{equation}
where $[\mu_{t}, \sigma_{t}] = \psi^{\text{Prior}}(z_{<t}^{Ns})$. The prior network is implemented using a spatio-temporal neural architecture. The complete prior for $z$ is given by:
\begin{equation}
	p(z_{t}) = p(z^{S}) \cdot \prod_{t=1}^{T} p(z_{t}^{Ns} | z_{<t}^{Ns}).
\end{equation}

\paragraph{Generative Model.}
The generation of $x_t$ at each time step is conditioned on both $z^{S}$ and $z_{t}^{Ns}$. The generative distribution is defined as:
\begin{equation}
	x_{t} | z^{S}, z_{t}^{Ns} \sim \mathcal{N}\left( \mu_{x,t}, \mathrm{diag}(\sigma_{x,t}^2) \right),
\end{equation}
where $[\mu_{x,t}, \sigma_{x,t}] = \theta^{\text{Decoder}}(z^{S}, z_{t}^{Ns})$. The decoder is implemented as a transposed convolutional network, reconstructing the sequence step by step. The joint distribution is:
\begin{equation}
	\begin{aligned}
		p_\theta(x_{1:T}, z_{1:T}^{Ns}, z^{S}) =\; & p_\theta(z^{S}) \cdot \\
		& \prod_{t=1}^{T} p_\theta(x_{t} | z^{S}, z_{t}^{Ns}) \cdot p_\theta(z_{t}^{Ns} | z_{<t}^{Ns}).
	\end{aligned}
\end{equation}

\paragraph{Inference Model.}
To approximate the posterior distribution, we use a deep structured encoder that factorises $z$ into disentangled groups. The amortised variational distribution is:
\begin{equation}
	q_\phi(z_{1:T}^{Ns}, z^{S} | x_{1:T}) = q_\phi(z^{S} | x_{1:T}) \cdot \prod_{t=1}^{T} q_\phi(z_{t}^{Ns} | x_{\leq t}).
\end{equation}

\paragraph{VAE Objective.}
The objective function is derived from the evidence lower bound (ELBO) of the standard VAE:
\begin{align}
	\mathcal{L}_{\text{VAE}}(\theta, \phi; x_{1:T}) = & \mathbb{E}_{q_\phi(z_{1:T}^{Ns}, z^{S} | x_{1:T})} \left[ \sum_{t=1}^{T} \log p_\theta(x_t | z_t^{Ns}, z^{S}) \right] \nonumber \\
	& - D_{\mathrm{KL}}(q_\phi(z^{S} | x_{1:T}) \,\|\, p_\theta(z^{S})) \nonumber \\
	& - \sum_{t=1}^{T} D_{\mathrm{KL}}(q_\phi(z_t^{Ns} | x_{\leq t}) \,\|\, p_\theta(z_t^{Ns} | z_{<t}^{Ns})).
	\label{eq:vae_loss_ns}
\end{align}

To further encourage disentanglement, we incorporate the total correlation estimation strategy from FactorVAE \citep{factorVAE}, using the independence testing trick and density-ratio estimation to approximate the KL term. This allows us to constrain the learned representations and separate sensitive from non-sensitive components more effectively.

This completes the generative modelling strategy in our framework. In the next section, we define the final loss function by incorporating additional regularisers that promote fairness and control representation leakage.

\subsection{Final Objective Function}
\label{sec:final_objective}

We encourage the representation $z^{S}$ to encode the most relevant information about the sensitive attributes $S$, while ensuring that $z^{Ns}$ contains minimal, if any, such information. The complete objective function of our framework is formulated as follows:
\begin{align}
	\mathcal{L}_{\text{FairDRL-ST}} & = \mathcal{L}_{\text{VAE}} + \lambda \left( \mathcal{L}_{S} - \mathcal{L}_{Ns} \right) \nonumber\\
	& = \mathcal{L}_{\text{VAE}} + \lambda \left( D_{f}( p_{S} \,\|\, p_{\hat{S}} ) - \frac{1}{n} \sum_{i=1}^{n} \left| P(z^{Ns}_i) - S_i \right| \right),
	\label{eq:final_loss}
\end{align}
where $\mathcal{L}_{S}$ measures the discrepancy between the distribution of the ground-truth sensitive attributes $S$ and the generated attributes $\hat{S}$, quantified using a divergence metric such as the KL divergence.

The second regulariser, $\mathcal{L}_{Ns}$, penalises the model if the non-sensitive representation $z^{Ns}$ retains predictive power over the sensitive attributes. This term is subtracted from the total loss, as we seek to minimise its value. The hyper-parameter $\lambda$ controls the trade-off between fairness and predictive utility, allowing us to adjust the strength of the disentanglement constraint depending on the desired balance.

\subsection{Fairness Evaluation Metrics}
\label{sec:fairness_metrics}

Although our framework does not incorporate fairness constraints directly into the training objective, evaluating the fairness of its predictions is essential for understanding their real-world impact. To achieve this, we adopt two fairness evaluation metrics: the Region-based Fairness Gap and the Individual-based Fairness Gap, both originally proposed by~\citet{yan2020fairness}. These metrics are not involved in the optimisation process but are applied after model training to assess whether the disentangled representations effectively reduce the influence of sensitive attributes on the predicted outcomes. We describe the definitions and motivations of these two metrics in the following.

\subsubsection{Region-based Fairness Gap (RFG)} To assess the fairness gap between advantaged and disadvantaged groups, the first step is to divide the cells based on a single sensitive attribute $S$ (e.g., race). Each cell $c_i$ in the study area is assigned a label, either $G^{+}$ (advantaged group) or $G^{-}$ (disadvantaged group), depending on the attributes of the majority of people living in that cell. 

The RFG between the two demographic groups over a period of time $T$ is defined as:
\begin{equation}
	\label{eq:rfg}
	RFG=\frac{\sum _{i\in G^{+}} E_{T}[\hat{y}_{i,t}]}{\sum _{i\in G^{+}} p_{i}} -\frac{\sum _{j\in G^{-}} E_{T}[\hat{y}_{j,t}]}{\sum _{j\in G^{-}} p_{j}}.
\end{equation}

The first term in Equation~\ref{eq:rfg} represents the per capita demand for group $G^{+}$, averaged over the time period $T$, while the second term represents the per capita demand for the disadvantaged group $G^{-}$ over the same period.

\subsubsection{Individual-based Fairness Gap (IFG)} While the RFG provides a broad measure of the fairness gap between two demographic groups, it can be inaccurate in certain circumstances. For instance, if 45\% of the people living in cell $c_i$ belong to the disadvantaged group, the cell would still be labelled as advantaged despite the small margin (10\%) between the two groups. The IFG is introduced to provide a more precise measurement of the gap between the two demographic groups in such cases.

Let $w^{+}$ and $w^{-}$ represent the percentage of people in the advantaged and disadvantaged groups, respectively, with respect to the sensitive attribute $S$. 

The IFG between the two demographic groups over a period of time $T$ is then defined as:
\begin{equation}
	\label{eq:ifg}
	IFG=\frac{\sum _{i\in S} E_{T}[\hat{y}_{i,t}] w_{i}^{+}}{\sum _{i\in S} p_{i} w_{i}^{+}} -\frac{\sum _{i\in S} E_{T}[\hat{y}_{i,t}] w_{i}^{-}}{\sum _{i\in S} p_{i} w_{i}^{-}}.
\end{equation}

The IFG assumes that the predicted mobility demand is proportionally allocated to all individuals in a cell, regardless of their demographic group. Therefore, the first term in Equation~\ref{eq:ifg} represents the average demand allocated to the advantaged group over the time period $T$, while the second term follows the same pattern for the disadvantaged group.

\section{Experiments}
\label{Experiments}

In this section, we empirically evaluate the effectiveness of the proposed framework, FairDRL-ST, in producing fair and accurate predictions on real-world spatio-temporal mobility datasets. Our evaluation is guided by the following research questions:

\begin{itemize}[leftmargin=0.6cm]
	\item \textbf{RQ1:} To what extent does our unsupervised FairDRL-ST reduce bias in the input data and generate fairness-aware predictions while maintaining high predictive accuracy?
	\item \textbf{RQ2:} How does the FairDRL-ST perform under varying levels of fairness constraints?
	\item \textbf{RQ3:} What is the relative contribution of each module in the proposed architecture to overall fairness and accuracy?
\end{itemize}

To answer these questions, we conduct comprehensive experiments on two real-world urban mobility datasets, TaxiNYC and BikeNYC. We compare FairDRL-ST against both supervised and unsupervised baselines, assess performance using a combination of accuracy and fairness metrics, and perform ablation studies to analyse the impact of individual components.

\begin{table}[t]
	\centering
	\caption{Overview of the 1D, 2D, and 3D datasets used in this study. Temporal datasets represent 1D inputs, spatial datasets represent 2D inputs, and spatio-temporal datasets correspond to 3D inputs.}
	\label{table:features}
	\begin{tabular}{ccc}
		\toprule
		Name & Type & Source \\
		\midrule 
		Temperature \  & Temporal & NCEI \\
		Pressure & Temporal & NCEI \\
		Humidity & Temporal & NCEI \\
		POI (business) & Spatial & NYC Open Data \\
		POI (food) & Spatial & NYC Open Data \\
		POI (government) & Spatial & NYC Open Data \\
		POI (hospitals) & Spatial & NYC Open Data \\
		POI (schools) & Spatial & NYC Open Data \\
		POI (transportation) & Spatial & NYC Open Data \\\midrule
		BikeNYC & Spatio-temporal & \citet{lin2019deepstn+} \\
		TaxiNYC & Spatio-temporal & \citet{yao2019revisiting} \\
		\bottomrule
	\end{tabular}
\end{table}

\subsection{Datasets}
In this work, we focus on the fairness and accuracy of the predictions made by our proposed model on spatio-temporal datasets. We use the following real-world urban flow datasets:

\begin{itemize}[leftmargin=0.5cm]
	\item \textbf{TaxiNYC} \citep{yao2019revisiting} is a dataset containing taxi in-out flow data from the New York City taxi system, created from NYC-Taxi GPS data, covering the period from 2015-01-01 to 2015-03-01. The spatio-temporal raster data is generated by integrating taxi data collected every 30 minutes.
	
	\item \textbf{BikeNYC} \citep{lin2019deepstn+} is a shared bike travel dataset that tracks the trajectory of all shared bikes provided by New York City's Citi Bike service. To convert this into a grid format, we calculate the in/out flow for each cell every 30 minutes, with the dataset covering the period from 2014-04-01 to 2014-09-30.
	
	\item \textbf{Other contextual datasets}: To improve prediction accuracy and enable the construction of the Equitensor \citep{yan2021equitensors}, we include several 1D and 2D datasets to extract contextual features. As shown in Table \ref{table:features}, the 1D features include temporal data describing the weather across New York City, collected by the National Centers for Environmental Information (NCEI) \citep{ncei}. The 2D features include points of interest (POI) data, which potentially influence mobility patterns. All contextual data is gathered from New York City agencies \citep{city_of_ny}.
\end{itemize}

%\begin{table*}[t]
%	\centering
%	\caption{The results of our proposed method compared to the baselines for TaxiNYC and BikeNYC predictions.}
%	\label{table:main_results}
%	\begin{tabular}{c|c|c|cccc|cccc}
%		\toprule 
%		&  & & \multicolumn{4}{c}{TaxiNYC} & \multicolumn{4}{c}{BikeNYC} \\
%		\midrule 
%		& Setting & $\displaystyle \lambda $ & MAE & RFG & IFG & SR & MAE & RFG & IFG & SR \\
%		\midrule 
%		Ground Truth & / & / & / & 114.763 & 42.935 & 0.103 & / & 79.84 & 55.43 & 0.125 \\
%		HA & unsupervised & / & 7.121 & 192.314 & 75.556 & 0.533 & 1.500 & 47.35 & 38.23 & 0.129 \\
%		CNN & supervised & / & 6.884 & 273.918 & 110.712 & 0.497 & 1.995 & 60.87 & 44.31 & 0.105 \\
%		ConvLSTM & supervised & / & 4.811 & 73.485 & 31.525 & 0.231 & 1.133 & 65.19 & 47.62 & 0.118 \\
%		FairST & supervised & 0 & 4.535 & 88.172 & 25.911 & 0.135 & 1.169 & 72.52 & 50.68 & 0.093 \\
%		FairST + RF & supervised & 0.05 & 4.721 & 79.570 & 24.694 & 0.067 & 1.185 & 50.38 & 40.95 & \textbf{-0.032} \\
%		FairST + IF & supervised & 0.60 & 4.853 & \textbf{10.627} & \textbf{3.363} & \textbf{-0.012} & 1.245 & 36.8 & \textbf{20.18} & -0.101 \\
%		Equitensor - 3 & unsupervised & 0.60 & 4.928 & 63.130 & 15.281 & 0.099 & 1.144 & 46.70 & 50.15 & 0.085 \\
%		Equitensor - 9 & unsupervised & 0.60 & \textbf{4.441} & 38.473 & 6.902 & 0.082 & 1.136 & 40.35 & 22.56 & 0.074 \\
%		\midrule
%		FairDRL-ST & unsupervised & 0.60 & 4.474 & 24.451 & 4.893 & 0.060 & \textbf{1.128} & \textbf{32.16} & 23.28 & 0.052 \\
%		\bottomrule
%	\end{tabular}
%\end{table*}

\begin{table*}[t]
	\centering
	\caption{The results of our proposed FairDRL-ST compared to baseline methods on the TaxiNYC and BikeNYC datasets. {Equitensor - 3} refers to the Equitensor method using 3 context features, while {Equitensor - 9} uses 9 context features. A slash ``/'' indicates that the corresponding value is not applicable, such as the ground truth for MAE or methods without fairness constraints for $\lambda$. The best results are highlighted in bold, and the second-best are underlined.}
	\label{table:main_results}
	\begin{tabular}{c|c|c|cccc|cccc}
		\toprule 
		&  & & \multicolumn{4}{c}{TaxiNYC} & \multicolumn{4}{|c}{BikeNYC} \\
		\midrule 
		& Setting & $\lambda$ & MAE ($\downarrow$) & RFG ($\downarrow$) & IFG ($\downarrow$) & SR ($\rightarrow0$) & MAE ($\downarrow$) & RFG ($\downarrow$) & IFG ($\downarrow$) & SR ($\rightarrow0$) \\
		\midrule 
		Ground Truth & / & / & / & 114.76 & 42.94 & 0.10 & / & 79.84 & 55.43 & 0.13 \\
		HA & / & / & 7.12 & 192.31 & 75.56 & 0.53 & 1.50 & 47.35 & 38.23 & 0.13 \\
		CNN & supervised & / & 6.88 & 273.92 & 110.71 & 0.50 & 2.00 & 60.87 & 44.31 & 0.11 \\
		ConvLSTM & supervised & / & {4.81} & {73.49} & {31.53} & {0.23} & \underline{1.13} & 65.19 & 47.62 & 0.12 \\
		FairST & supervised & 0.00 & 4.54 & 88.17 & 25.91 & 0.14 & 1.17 & 72.52 & 50.68 & 0.09 \\
		FairST + RF & supervised & 0.05 & 4.72 & 79.57 & 24.69 & 0.07 & 1.19 & 50.38 & 40.95 & \textbf{-0.03} \\
		FairST + IF & supervised & 0.60 & 4.85 & \textbf{10.63} & \textbf{3.36} & \textbf{-0.01} & 1.25 & \underline{36.80} & \textbf{20.18} & -0.10 \\\midrule
		Equitensor - 3 & unsupervised & 0.60 & 4.93 & 63.13 & 15.28 & 0.10 & 1.14 & 46.70 & 50.15 & 0.09 \\
		Equitensor - 9 & unsupervised & 0.60 & \textbf{4.44} & 38.47 & 6.90 & 0.08 & 1.14 & 40.35 & \underline{22.56} & {0.07} \\
		\midrule
		FairDRL-ST & unsupervised & 0.60 & \underline{4.47} & \underline{24.45} & \underline{4.89} & \underline{0.06} & \textbf{1.13} & \textbf{32.16} & 23.28 & \underline{0.05} \\
		\bottomrule
	\end{tabular}
\end{table*}

\begin{figure*}[t]
	\begin{center}
		\includegraphics[width=0.99\textwidth]{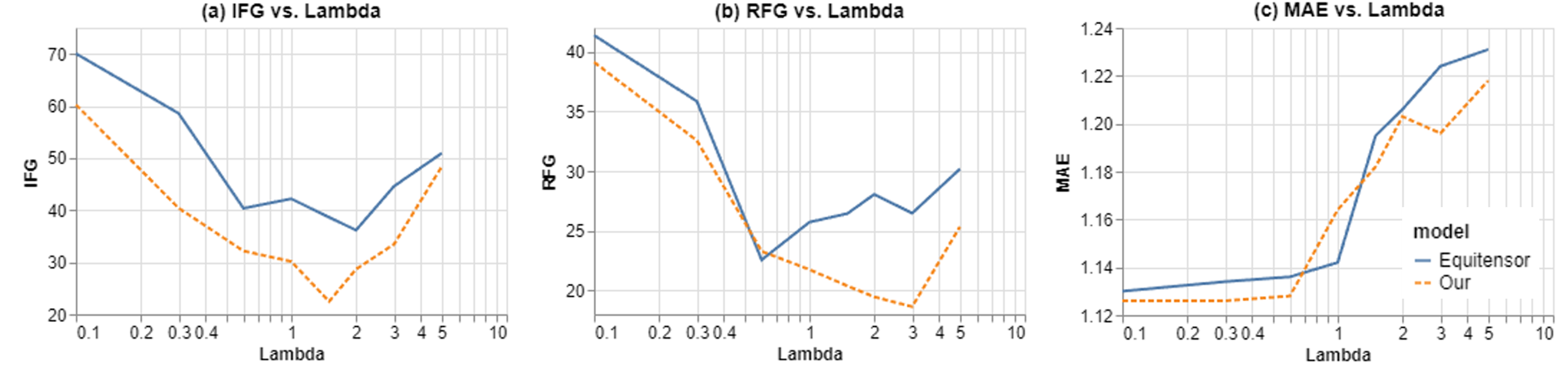}
	\end{center}
	\caption{Figures (a), (b), and (c) illustrate the relationship between the regularisation weight $\lambda$ and the fairness metrics RFG and IFG, as well as the prediction accuracy metric MAE, on the BikeNYC dataset. Lower values on the y-axis indicate better performance.}
	\label{fig:changing_lambda}
\end{figure*}

\subsection{Baseline Methods}
To evaluate the performance of our proposed framework, we compare it with the following baseline methods:
\begin{itemize}[leftmargin=0.5cm]
	\item \textbf{Historical Average (HA)}: A simple yet widely used baseline for mobility forecasting tasks, which predicts future values by averaging historical observations.
	
	\item \textbf{Convolutional Neural Network (CNN)}: A standard deep learning method that captures local spatial patterns in mobility data through convolutional operations.
	
	\item \textbf{Convolutional LSTM (ConvLSTM)}~\citep{shi2015convolutional}: An extension of the fully connected LSTM (FC-LSTM) that incorporates convolutional structures to jointly capture spatial and temporal dependencies. It has shown promising performance in tasks such as mobility prediction and video modelling.
	
	\item \textbf{Fairness-aware Spatial-Temporal (FairST)}~\citep{yan2020fairness}: A supervised fairness-aware method that integrates fairness metrics into the learning objective to reduce disparities in mobility predictions. Two variants are considered: {FairST+RF}, which incorporates the RFG, and {FairST+IF}, which incorporates the IFG. Both aim to produce fairness-aware predictions with minimal degradation in accuracy.

	\item \textbf{Equitensor}~\citep{yan2021equitensors}: An unsupervised spatio-temporal method designed to learn debiased representations from multiple datasets. It uses an adaptive weighting scheme to balance gradients across datasets, effectively mitigating demographic bias and enabling fair mobility forecasting.
\end{itemize}

\subsection{Evaluation Metrics}
We evaluate method performance using both accuracy and fairness metrics. For prediction accuracy, we report Mean Absolute Error (MAE). To assess fairness, we use RFG and IFG, as discussed in Section~\ref{sec:fairness_metrics}. 

Additionally, we consider Spearman’s rank correlation coefficient (SR) to measure the correlation between predicted demand rankings and the proportion of the disadvantaged group across regions. This metric helps evaluate whether high-demand predictions disproportionately favour advantaged neighbourhoods. An SR close to zero is ideal, as it reflects independence between predictions and demographic composition. However, excessively large negative correlations may indicate overcompensation or introduce reverse bias.
 
\subsection{RQ1: Main Results}
We begin by examining whether our proposed unsupervised FairDRL-ST can effectively reduce bias in the input data and generate fairness-aware predictions without significantly compromising accuracy. This question is central to the goal of equitable mobility resource allocation, which requires not only fair outcomes but also reliable prediction of underlying mobility patterns. A well-performing model should therefore minimise discrepancies between predicted and actual demand, while also addressing fairness concerns across demographic groups.

The results for both real-world datasets are reported in Table~\ref{table:main_results}. Although the hyper-parameter $\lambda$ is present in FairST, Equitensor, and our method, its role varies across them. In FairST, a supervised method, $\lambda$ directly controls the trade-off between accuracy and fairness by incorporating fairness metrics such as RFG and IFG into the training objective. In contrast, Equitensor and our method adopt an unsupervised setting, where $\lambda$ regulates the strength of constraints designed to prevent sensitive information from being encoded in the non-sensitive representation. While fairness is not explicitly optimised in these methods, increasing $\lambda$ still leads to fairness improvements, with a more stable impact on predictive performance.

The proposed FairDRL-ST achieves the best prediction accuracy on the BikeNYC dataset. More importantly, across both datasets, it consistently reduces fairness gaps while maintaining competitive accuracy. Although FairST achieves lower fairness gaps in some settings, its performance declines rapidly under stronger fairness constraints. This highlights a key difference: our unsupervised method is not explicitly trained on fairness metrics, yet it still achieves a favourable balance between fairness and accuracy, demonstrating robustness and strong generalisation.

\begin{table*}[t]
	\centering
	\caption{Ablation study on the TaxiNYC and BikeNYC datasets. The best results are highlighted in bold.}
	\label{table:ablation_results}
	\begin{tabular}{cc|c|ccc|ccc}
		\toprule
		\multicolumn{2}{c|}{Setup} &  & \multicolumn{3}{c}{TaxiNYC} & \multicolumn{3}{|c}{BikeNYC} \\
		\midrule 
		Sensitive Module & Non-sensitive Module & $\displaystyle \lambda $ & MAE ($\downarrow$) & RFG ($\downarrow$) & IFG ($\downarrow$) & MAE ($\downarrow$) & RFG ($\downarrow$) & IFG ($\downarrow$) \\
		\midrule 
		\ding{51} & \ding{55} & 0.1 & 4.79 & 68.14 & 16.27 & 1.14 & 65.88 & 51.34 \\ 
		\ding{51} & \ding{55} & 0.3 & \textbf{4.36} & 45.04 & 9.27 & 1.49 & 50.16 & 44.37 \\ 
		\ding{51} & \ding{55} & 0.6 & 4.85 & 34.57 & 6.39 & 1.14 & 38.67 & 26.83 \\
		\ding{55} & \ding{51} & 0.1 & 4.46 & 50.12 & 13.13 & 1.13 & 55.21 & 50.95 \\
		\ding{55} & \ding{51} & 0.3 & 4.51 & 38.93 & 6.99 & 1.20 & 48.87 & 40.03 \\
		\ding{55} & \ding{51} & 0.6 & 4.51 & 32.47 & 6.17 & 1.29 & 34.32 & 24.21 \\
		\ding{51} & \ding{51} & 0.1 & 4.47 & 40.26 & 7.03 & \textbf{1.13} & 60.24 & 39.12 \\
		\ding{51} & \ding{51} & 0.3 & 4.48 & 27.13 & 5.89 & \textbf{1.13} & 40.38 & 32.56 \\
		\ding{51} & \ding{51} & 0.6 & 4.47 & \textbf{24.45} & \textbf{4.89} & 1.128 & \textbf{32.16} & \textbf{23.28} \\
		\bottomrule
	\end{tabular}
\end{table*}

\subsection{RQ2: Impact of Constraint Strength on Fairness-Utility Trade-off}
We further investigate how our framework behaves under different levels of constraint imposed by the fairness-aware regularisation. Unlike FairST, which is trained in a supervised manner with fairness objectives directly embedded into the loss function, our method is unsupervised and does not rely on fairness labels. Therefore, increasing the strength of the constraint may not always lead to direct improvements in fairness metrics.

To evaluate this, we vary the regularisation weight $\lambda$ and compare the results against Equitensor, which is the only other unsupervised method designed for fairness in spatio-temporal prediction. The results are presented in Figure~\ref{fig:changing_lambda}.

\begin{figure}[t]
	\begin{center}
		\includegraphics[width=0.33\textwidth]{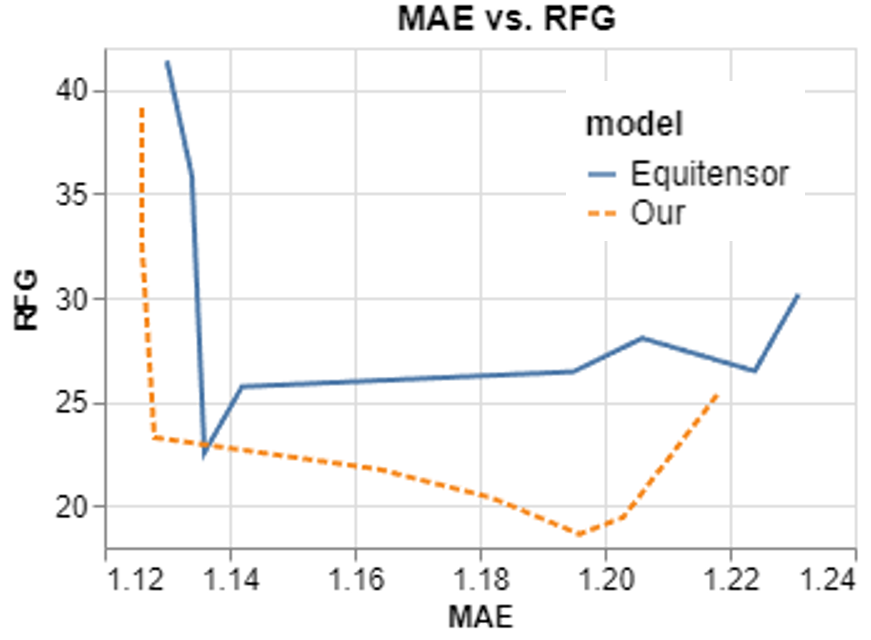}
	\end{center}
	\caption{The trade-off between MAE and RFG for both Equitensor and our proposed FairDRL-ST. Lower values on the vertical axis indicate better overall performance in terms of fairness and accuracy.}
	\label{fig:(d)}
\end{figure}

Figures~\ref{fig:changing_lambda}(a), (b), and (c) show the effects of $\lambda$ on RFG, IFG, and MAE, respectively. Across most settings, our framework outperforms Equitensor, especially when both models are calibrated to yield comparable predictive performance. Notably, our framework maintains stronger fairness performance in cases where Equitensor begins to overfit. This indicates that our framework achieves a more stable trade-off between fairness and utility in mobility forecasting tasks. Figure~\ref{fig:(d)} further supports this observation, where the performance curve of our method lies consistently closer to the lower-left corner, indicating jointly lower error and fairness gaps.

However, we also observe that the performance of both methods degrades when the constraint becomes too strong. This is likely due to the fairness regularisation overpowering the method's ability to retain task-relevant information, resulting in diminished accuracy. Additionally, fairness degradation occurs earlier in the IFG metric compared to RFG. This may be attributed to IFG’s sensitivity to group proportions within each spatial unit, making it more vulnerable to information loss under strong constraint regimes.

\subsection{Ablation Study}
To evaluate the effectiveness of different components within our framework, we conduct an ablation study. Specifically, we examine how the exclusion of either the \textit{sensitive module} or the \textit{non-sensitive module} affects predictive performance and fairness outcomes. In each ablation setting, the removed module is kept fixed throughout training, and its corresponding loss is set to zero. The results for both TaxiNYC and BikeNYC are presented in Table~\ref{table:ablation_results}.

We observe that removing either module generally leads to higher fairness gaps, as measured by both RFG and IFG. This finding confirms that both modules play an important role in promoting fairness. Among them, the \textit{non-sensitive module} appears to contribute more significantly to reducing disparities between demographic groups. For example, settings that include only the non-sensitive module tend to yield better fairness performance compared to those that include only the sensitive module. However, a different pattern is observed when we evaluate MAE. On the TaxiNYC dataset, the best prediction accuracy is achieved when only the sensitive module is active and the regularisation parameter $\lambda$ is set to 0.3. A similar result is found on the BikeNYC dataset, where this configuration also produces low MAE. In contrast, including the non-sensitive module in these cases leads to slightly higher prediction error. This is likely because the removal of sensitive information also reduces access to features that are still useful for prediction.

In summary, the ablation study reveals a trade-off between fairness and accuracy. While the complete model with both modules consistently achieves the lowest fairness gaps, it does not always produce the best prediction performance. These findings highlight the challenge of achieving fairness in unsupervised representation learning, where sensitive attributes must be suppressed without compromising the utility of the learned features.

\section{Conclusion}
\label{Conclusion}
In this paper, we present FairDRL-ST, a fairness-aware and unsupervised spatio-temporal prediction framework designed for mobility demand forecasting. The framework integrates adversarial learning with disentangled representation learning to separate sensitive attributes from task-relevant features, enabling fair representation learning without relying on demographic group labels during training. Extensive experiments on real-world urban mobility datasets demonstrate that FairDRL-ST consistently improves fairness outcomes across individual and regional metrics while maintaining or enhancing predictive accuracy compared to both supervised and unsupervised baselines. Additional analyses, including the impact of varying constraint strengths and an ablation study, further confirm the stability and effectiveness of the proposed framework. This work contributes to the broader effort of promoting fairness in machine learning by addressing bias in complex spatio-temporal domains. Future work will explore adaptive constraint mechanisms to better balance fairness and accuracy, and extend the framework to other applications such as traffic flow forecasting and urban resource allocation.

\bibliographystyle{ACM-Reference-Format}
\bibliography{mybib}

\appendix

\end{document}